\documentclass[sigconf]{acmart}

\AtBeginDocument{%
  }

\copyrightyear{2022}
\acmYear{2022}
\setcopyright{rightsretained}
\acmConference[SA '22 Conference Papers]{SIGGRAPH Asia 2022 Conference
Papers}{December 6--9, 2022}{Daegu, Republic of Korea}
\acmBooktitle{SIGGRAPH Asia 2022 Conference Papers (SA '22 Conference
Papers), December 6--9, 2022, Daegu, Republic of
Korea}\acmDOI{10.1145/3550469.3555421}
\acmISBN{978-1-4503-9470-3/22/12}

\usepackage{subcaption}
\usepackage{graphbox}
\usepackage{wrapfig}
\usepackage{comment}

\begin{document}
\bibliographystyle{ACM-Reference-Format}
\title{Physical Interaction: Reconstructing Hand-object Interactions with Physics}

\author{Haoyu Hu}
\email{hhythu17@163.com}
\affiliation{%
  \institution{School of software and BNRist, Tsinghua University}
  \country{China}
}

\author{Xinyu Yi}
\email{yixy20@mails.tsinghua.edu.cn}
\affiliation{%
  \institution{School of software and BNRist, Tsinghua University}
  \country{China}
}

\author{Hao Zhang}
\email{zhanghao_buaa@163.com}
\affiliation{%
  \institution{Institute of Computer Application, China Academy of Engineering Physics (CAEP)}
  \country{China}
}

\author{Jun-Hai Yong}
\email{yongjh@tsinghua.edu.cn}
\affiliation{%
  \institution{School of software and BNRist, Tsinghua University}
  \country{China}
}

\author{Feng Xu}
\email{xufeng2003@gmail.com}
\affiliation{%
  \institution{School of software and BNRist, Tsinghua University}
  \country{China}
}


\begin{abstract}
Single view-based reconstruction of hand-object interaction is challenging due to the severe observation missing caused by occlusions. 
This paper proposes a physics-based method to better solve the ambiguities in the reconstruction.
It first proposes a force-based dynamic model of the in-hand object, which not only recovers the unobserved contacts but also solves for plausible contact forces. 
Next, a confidence-based slide prevention scheme is proposed, which combines both the kinematic confidences and the contact forces to jointly model static and sliding contact motion. 
Qualitative and quantitative experiments show that the proposed technique reconstructs both physically plausible and more accurate hand-object interaction and estimates plausible contact forces in real-time with a single RGBD sensor.

\end{abstract}



\maketitle

\section{Introduction}
\begin{figure}[t]
\includegraphics [width=0.94\linewidth] {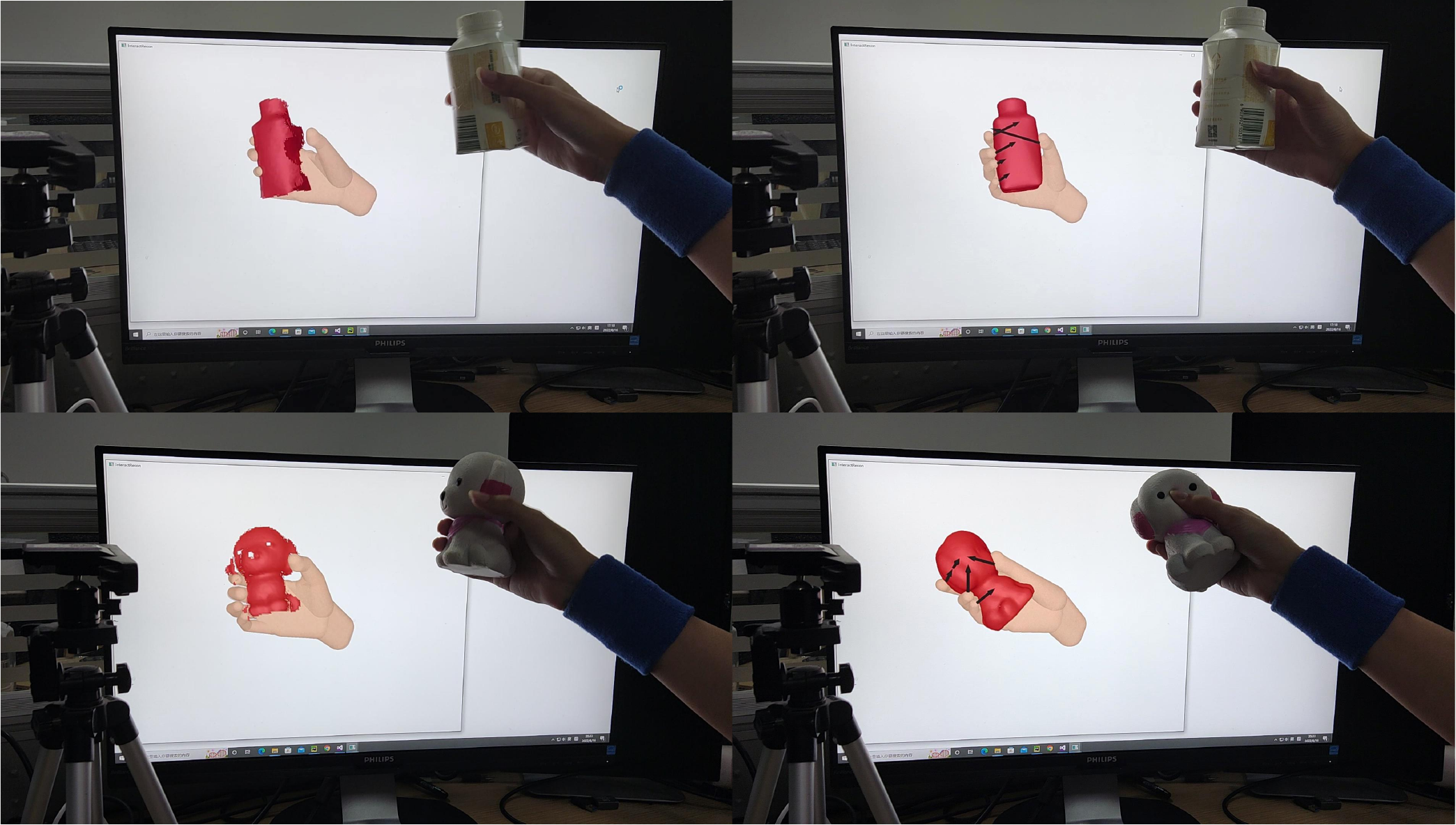} 
\caption{Our system reconstructs physically plausible hand-object interaction and contact forces in real-time from a single-view depth camera.}
\label{fig:teaser}
\end{figure}

In daily lives, humans usually interact with objects with their hands.
Thus hand-object interaction is one important kind of motion to be handled in computer vision and graphics.
And reconstructing hand-object interactions is very useful in various applications including gaming, virtual reality, human-computer interaction, and robotics.
%
%
%
\par
%
%
%
%
%
%
%
Many previous works have worked on the task of reconstructing hand-object interactions.
Recently, the state-of-the-art work~\cite{zhang2021single} reconstructs both the hand and the in-hand object in real-time with only a single depth camera.
They achieve both promising reconstruction accuracy and runtime performance.
However, hand-object interaction usually causes heavy occlusions, leading to severe observation missing, especially in the single view reconstruction scenario.
\cite{zhang2021single} rely on data-driven pose priors to handle this.
However, they cannot fully solve the ambiguities, and may fail or generate noticeable artifacts, especially some physically implausible failures and artifacts, \textit{e.g.,} grasping an object with only one finger in contact, or the object slides in the hand, making the technique difficult to be used in many real applications.
%
%
\par
These problems motivate us to not only utilize the data-driven priors but also involve physics-driven priors to better solve the ambiguities in the reconstruction.
However, it is not trivial to build physical models for hand-object interactions.
In body motion capture, we know that physics-based contact and sliding optimizations have been successfully explored~\cite{PhysCap,PIP,PhysAware} and are demonstrated useful for realistic human motion estimation. 
However, different from foot-floor contacts, hand-object contacts are more complicated due to the high freedom of finger motions and the diversity of object shapes. 
Specifically, \textit{1)} hand-object contacts are difficult to detect because of severe occlusions. 
\textit{2)} And modeling the sliding constraints of hand contacts is difficult since fingers can either slide or fix on an object during the interaction.
\par
%
In this paper, we focus on building a physical model for reconstructing hand-object interactions. 
We observe that previous works do not delicately model the physics of hand-object interactions.
The state-of-the-art work~\cite{zhang2021single} just models the interaction kinematically with only straightforward sliding and penetration constraints, which lacks the awareness of interaction physics.
%
%
%
%
%
On the other hand, we incorporate delicately designed physics into the traditional kinematic reconstruction of hand-object interaction.
First, based on the physical rule that the object's motion is driven by the forces exerted at the contact points, we propose a novel contact status optimization to iteratively refine the contact status and estimate the contact forces to leverage the object's dynamics.
In this manner, some missing contacts, caused by occlusions, can be recovered by enforcing the consistency between the object motion and the forces on the contacts.   
Second, with the force estimation, we aim to examine the occurrence of \textit{non-physical sliding} at the contacts, \textit{i.e.,} the contact point falsely slides on the object surface when there is a large pressure on the contact.
To deal with false positives caused by the uncertainty in force (pressure) estimation, we design a novel confidence-based slide prevention method, which considers both kinematic confidences and contact forces.
With these considerations, we achieve reasonable contact motion estimation, handling both static and sliding contacts.
%
%
We demonstrate that by incorporating interaction physics into the reconstruction system, we achieve not only physical plausible results but also more accurate hand pose estimation, because the physical priors help to better solve the ambiguities caused by occlusions and noise in the input.
\par
%
%
%
%
%
%
%
%
%
%
In summary, our main contributions are:
\begin{itemize}
    \item The first method that reconstructs interacting hand and object with physically plausible contact motions and forces in real-time.
    \item A physics-based contact status optimization algorithm, which iteratively refines the hand-object contacts leveraging the object dynamics.
    \item A confidence-based slide prevention algorithm, which eliminates non-physical sliding by incorporating both kinematic confidence and physical force estimation.
\end{itemize}
%

\section{Related Works}
This paper studies physics-inspired algorithms to improve hand-object interaction reconstruction.
So, we majorly review the topics of hand-object interaction reconstruction and physics-based dynamic reconstruction.  
Notice that there exist numerous methods that focus on pure hand tracking or pure object reconstruction from interacting motions. As they do not simultaneously reconstruct both of them as we do, we will not provide a thorough discussion on them.
%
%
%
\subsection{Hand-object Interaction Reconstruction}
There are various works aiming at jointly tracking hand pose and object motion during the interaction. 
Many of them utilize optimization-based methods to find the solutions that best fit the observations which are obtained from different sorts of input.
\cite{oikonomidis2011full,wang2013video,ballan2012motion} reconstruct hand-object interactions by multi-view RGB cameras, while \cite{panteleris2017back} perform the reconstruction based on stereo RGB cameras. 
In recent years, depth cameras are widely applied because they can directly provide 3D information. 
\cite{kyriazis2014scalable} use RGBD data for motion tracking in complex scenes where the hand interacts with several objects, but the system can only run offline. 
Real-time performance is achieved by \cite{sridhar2016real} with RGBD input and a 3D articulated Gaussian mixture alignment approach. 
All these works need templates of objects in motion estimation, which greatly limits their usage.
\cite{BMVC2015_123} track object rigid motions and progressively reconstruct the object shape without a template. 
%
Recent works \cite{zhang2019interactionfusion,zhang2021single} perform real-time reconstruction of articulated hand pose, object shape, and object rigid/non-rigid motion. 
Even though they involve data-driven priors to solve the severe ambiguities in the reconstruction, they still suffer from some challenging poses with heavy occlusions.
This paper focuses on further providing physics-driven priors to better solve the ambiguity. 

Apart from the optimization-based methods, recent works utilize neural networks to extract interaction information directly from single color images. 
\cite{tekin2019h+} propose a unified network which can simultaneously predict hand-object poses, object categories, and action classes. 
\cite{hasson2019learning} recover hand and object shapes together with motions through two separate networks. 
However, for techniques in this category, the generalization ability for novel object geometries is limited because it is difficult to make coverage of the training dataset sufficient, compared to the huge geometry variations in the real world.

\subsection{Physics-based Dynamic Reconstruction}
%
%
Recently, many researches have focused on physics-based dynamic reconstruction.
Some works studied physically plausible body motion capture~\cite{PhysCap,Li2019,Rempe2020,PIP,Zell2020,PhysAware}, character animation and control~\cite{DeepMimic,SFV,Yuan2019,Isogawa2020,SimPoE,Yu2021}. 
%
%
%
%
%
Many works focus on physics-based hand motion and hand-object interaction, which are more relevant to our topic.
%
%
%
%
%
Some works use trajectory optimization \cite{liu2009dextrous} or data-driven physical controller \cite{pollard2005physically, zhao2013robust} to synthesize physics-based dexterous manipulations.
Most recent works generate physics-based hand control policy from deep reinforcement learning to reach specific grasping or moving goals \cite{yang2022learning, christen2022d}.
Researches on virtual reality obtain realistic interaction motion by leveraging soft contact modeling and physics simulation \cite{talvas2015aggregate, hirota2016interaction, holl2018efficient}.
However, these works assume a known virtual object shape and aim at \textit{synthesizing} hand motion to manipulate the object, rather than hand-object \textit{reconstruction}.
There are also works focusing on hand-object contact force estimation from visual inputs~\cite{pham2017hand,pham2015towards,ehsani2020use}.
They need a good hand-object motion estimation and aim at estimating the real forces during interaction, while we focus on improving the tracking accuracy leveraging physics.
Some works explore capturing or refining the hand-object interactive motion leveraging physics~\cite{grady2021contactopt,kumar2021physically,tzionas2016capturing,kry2006interaction}, which is the most relevant to ours.
However, these physical models either have strong assumptions like known object shapes or motion, or are not applicable for the real-time reconstruction task.
For the existing physical models of hand-object interaction, there is a gap between effectiveness to solve the ambiguity in the reconstruction task and efficiency to achieve real-time performance.

\begin{figure*}[!t]
	\centering
	\includegraphics[width=0.95\textwidth]{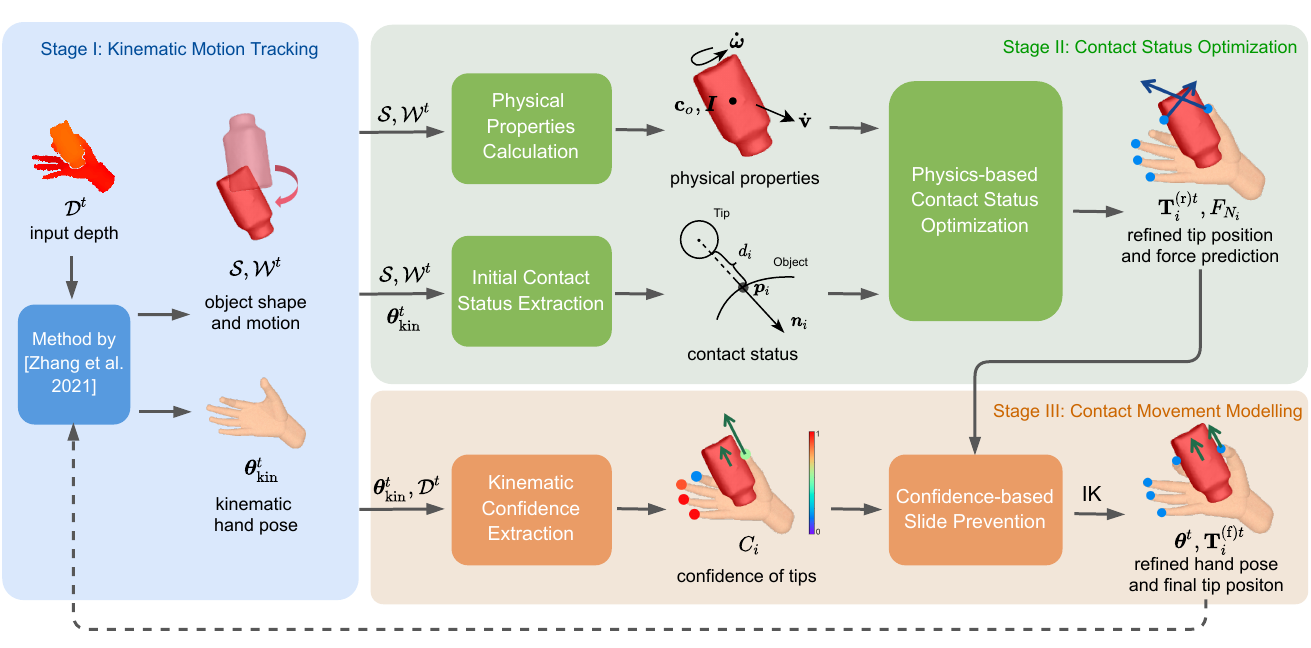}
	\caption{Method overview. We first track the hand-object motion using the method by \cite{zhang2021single}. Then, we refine the contact status based on the object dynamics. Finally, we model the hand-object movement at the contact points using a confidence-based algorithm. Our physics-based optimization ensures physical plausibility and improves accuracy.}
	\label{pipeline}
\end{figure*}

\section{Preliminary}
%
Similar to \cite{zhang2021single}, we use sphere-mesh~\cite{tkach2016sphere} to model the hand and Truncated Signed Distance Function (TSDF)~\cite{newcombe2015dynamicfusion} to model the object. 

Sphere-mesh model deems the human hand as a skeleton that consists of end spheres and connections between spheres. 
%
We describe the hand pose with a vector $\boldsymbol{\theta} \in \mathbb{R}^{28}$ that contains 6-DOF wrist pose and 22-DOF finger joint rotations.
The object is represented by a static model in zero pose in a canonical space and a motion field consisting of the object's rigid transformation and non-rigid deformation. 
The object's static model is defined as $\mathcal{S}=\{d(\boldsymbol{x}), w(\boldsymbol{x})\}$ where $\boldsymbol{x}$ is a point in the canonical space, $d(\boldsymbol{x})$ is the signed distance from $\boldsymbol{x}$ to the closest point on the object surface, and $w(\boldsymbol{x})$ measures the confidence of $d(\boldsymbol{x})$. 
The surface of the static object $\mathcal{M}$ is formulated as: 
\begin{equation}
    \mathcal{M}=\{(\boldsymbol{v, n})|d(\boldsymbol{v})=0,\boldsymbol{n}=\frac{\nabla{d(\boldsymbol{v})}}{||\nabla{d(\boldsymbol{v})||_2}}\},
\end{equation}
where $\boldsymbol v$ is the vertex and $\boldsymbol n$ is the corresponding normal.
The object motion is represented by a motion field $\mathcal{W}(\boldsymbol{x})$ which is the composition of a rigid transformation $\boldsymbol{W}$ and a local non-rigid deformation of each vertex. 
Applying the motion on the static object, we define the live model $\mathcal{M}_l$ as:
\begin{equation}
\mathcal{M}_l=\{(\boldsymbol{v}_l, \boldsymbol{n}_l)|\boldsymbol{v}_l=\mathcal{W}(\boldsymbol{v})\boldsymbol{v},\boldsymbol{n}_l=\mathcal{W}(\boldsymbol{v})\boldsymbol{n}\}.
\end{equation}
For more details, please refer to \cite{tkach2016sphere} and \cite{newcombe2015dynamicfusion}.
\section{Method}
In this section, we present the details of our method. 
%
%
Our task is to reconstruct the interacting hand and object in real-time from a single-view depth camera. 
The system obtains color and depth image input from the camera and outputs hand pose in terms of joint angles, object motion, and the object 3D model (Fig.~\ref{pipeline}).
Our method incorporates three stages: kinematic hand-object motion tracking (Sec.~\ref{sec:stage1}), physics-based contact status optimization (Sec.~\ref{sec:stage2}), and confidence-based contact movement modeling (Sec.~\ref{sec:stage3}).
At the beginning of the motion, we only run the kinematic motion tracking to estimate the object's static model. After the object is fully reconstructed, we calculate the object's physical properties and activate the physics-based optimization.
As the hand-object contacts can be very dense and complex, it is necessary to simplify the contacts to ensure a stable physics-based optimization and real-time performance. 
Thus, in this paper, we model the hand-object contacts only on \textit{fingertips} since they are the most common and important area for hand-object interaction, and suffer from non-physical artifacts (\textit{e.g.,} sliding, penetration, false escape from the object surface) most in previous works.
In the following, we elaborate each stage respectively.

\subsection{Kinematic Hand-Object Motion Tracking}\label{sec:stage1}
This stage estimates hand pose $\boldsymbol{\theta}^t_{\mathrm{kin}}$, object motion $\mathcal{W}^t$, and the static object model $\mathcal{S}$ from the input depth map $\mathcal{D}^t$ at frame $t$.
Notice that we use \cite{zhang2021single} for this kinematic reconstruction step, but in theory, it is applicable for other alternations as our technique is just a refining method working on top of a reconstruction system.
Also note that for some symbols that are always associated with frame $t$, we may eliminate $t$ for simplification.

\subsection{Contact Status Optimization}\label{sec:stage2}
The kinematic motion tracking in Sec.~\ref{sec:stage1} suffers from physically incorrect hand-object interaction. 
In this subsection, we focus on the most important and fundamental part of the interaction, \textit{i.e.,} contacts. 
Due to the insufficient depth data caused by the single view recording and the hand-object occlusions, we often observe physically implausible contacts in the kinematic tracking results, \textit{e.g.,} the object floats in the air without any finger supporting it. 
Our idea is to refine the contact status by leveraging the physical prior that the movement of the object should be explained by a group of forces exerted at all contact points.
In the following, we introduce the initial contact status extraction and the physics-based contact status refinement.


\subsubsection{Initial Contact Status Extraction}
As we assume contacts happen on the five fingertips. The contact status can be formulated as:
\begin{equation}\label{eq:cs}
    \mathcal{CS}=\{(\boldsymbol{p}_i, \boldsymbol{n}_i, d_i)|i=0,1,...4\},
\end{equation}
where $\boldsymbol{p}_i$ is the $i$th fingertip's projection point on the object surface, $\boldsymbol{n}_i$ is the normal of the object at $\boldsymbol{p}_i$, and $d_i$ is the distance between the tip and $\boldsymbol{p}_i$.
The projection point $\boldsymbol{p}_i$ can be treated as the candidate position of the $i$th fingertip's contact point on the object surface. 
$d_i$ somehow represents whether the $i$th fingertip is in contact with the object ($d_i=0$ means contact). 
Here we discuss how to obtain $\mathcal{CS}$.
First, we find the candidate contact on the five fingertips.
We sample some surface points on each of the fingertip, and based on their positions in the kinematic result, we examine their values in the TSDF of the reconstructed object.
The point with minimum values is treated as the candidate as it is the closest to the object. %
Then we project this point onto the object along the normal direction to get $\boldsymbol{p}_i$. 
And thus $\boldsymbol{n}_i$ and $d_i$ can also be calculated.
Notice that \cite{zhang2021single} also has a contact detection scheme. 
It is also applicable to use their method to construct $\mathcal{CS}$.

\subsubsection{Physical Properties Calculation}
To perform physics-based contact status optimization, we also need to compute the object's physical properties including dynamics (linear and angular velocity $\boldsymbol{v}^t, \boldsymbol{\omega}^t$, linear and angular acceleration $\dot{\boldsymbol{v}}^t, \dot{\boldsymbol{\omega}}^t$) and inherent attributes (mass $m_o$, center of mass $\boldsymbol{c}_o$, and inertia tensor $\boldsymbol{I}$). 
The dynamics are derived from previous and current object rigid motion $\{\boldsymbol{W}^{t-2}, \boldsymbol{W}^{t-1}, \boldsymbol{W}^{t}\}$ using the finite difference method.
For the inherent attributes, since mass and its distribution cannot be learned visually, we use a fixed mass $m_o = 0.2\mathrm{kg}$ and assume a hollow object, \textit{i.e., } mass is evenly distributed on the surface of the object. 
Note that $m_o$ is just a relative value for solving forces and does not affect the physical refinement.
%
%
Although not accurate, this simple assumption can work well in preventing the physically implausible failures and artifacts. 
The center of mass $\boldsymbol{c}_o$ is the mean position of all vertices on the surface and inertia $\boldsymbol{I}$ can be computed by:
\begin{align}
\boldsymbol{I}_{ij}=\frac{m_o}{|\mathcal{M}|}\sum_{k=1}^{|\mathcal{M}|}(||\boldsymbol{r}_k||^2\delta_{ij}-x_i^{(k)}x_j^{(k)}), \quad i,j\in\{1, 2, 3\} \nonumber \\ 
\boldsymbol{r}_k = \boldsymbol{v}_k - \boldsymbol{c}_o = (x_1^{(k)}, x_2^{(k)}, x_3^{(k)}), \quad (\boldsymbol{v}_k, \boldsymbol{n}_k) \in \mathcal{M},
\end{align}
where $\delta_{ij}$ is the Kronecker delta, and $\boldsymbol{I}_{ij}$ is the $(i, j)$ entry of the inertia tensor $\boldsymbol{I} \in \mathbb{R}^{3\times 3}$. 

\subsubsection{Physics-based Contact Status Optimization} 
This step takes the input of the initial contact status $\mathcal{CS}=\{(\boldsymbol{p}_i, \boldsymbol{n}_i, d_i)\}$, the object's rigid motion $\boldsymbol{W}^t$, and the physical properties $\boldsymbol{\omega}^t, \dot{\boldsymbol{v}}^t, \dot{\boldsymbol{\omega}}^t, m_o, \boldsymbol{c}_o,$
$\boldsymbol{I}$.
Our target is to optimize a set of new tip-object distances $\{ \tilde{d}_i \}$ which better satisfies the interaction physics.
To achieve this, we simultaneously refine the tip-object distances $\tilde{d}_i$ and estimate the force $\boldsymbol{F}_i$ exerted by each tip on the object to explain the object's motion.
The intuition behind is that: \textit{1)} to explain the object's motion, we must have a set of physically correct forces applied on the object, and \textit{2)} based on the condition that a fingertip must \textit{touch} the object to exert force on it, we obtain better contact states by force estimation.
Therefore, our optimizer is mathematically defined as:
\begin{align}
E(\boldsymbol{f}_i,\tilde{d}_i) = E_{\text{f}}(\boldsymbol{f}_i) + E_{\text{m}}(\boldsymbol{f}_i) + E_{\text{reg}}(\boldsymbol{f}_i)& + E_{\text{tac}}(\boldsymbol{f}_i,\tilde{d}_i) + E_{\text{smo}}(\tilde{d}_i),\nonumber \\ 
\text{s.t. } \boldsymbol{f}_i\ge0,\tilde{d}_i\ge0.
\label{mainEnergyOfStageII}
\end{align}
The notation $\boldsymbol{f}_i$ and each energy term is elaborated in the following.

\begin{figure}[H]
    \begin{minipage}[t]{0.45\linewidth}
        \centering
        \includegraphics[width=1.0\textwidth]{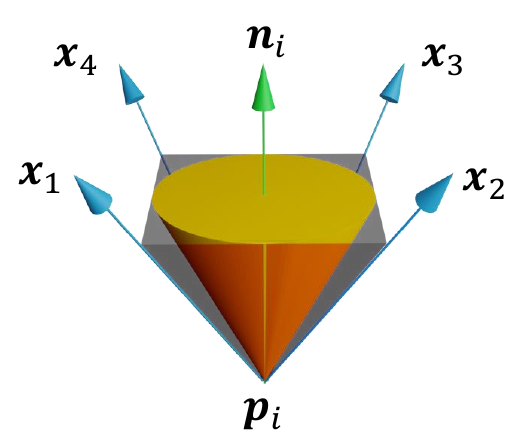}
    	\caption{Friction cone (yellow) and its approximation (gray). 
    	}
    	\label{fig:fri-cone}
    \end{minipage}
    \qquad
    \begin{minipage}[t]{0.45\linewidth}
        \centering
        \includegraphics[width=1.0\textwidth]{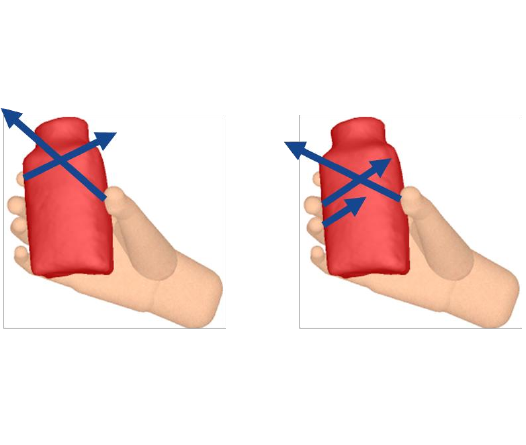}
    	\caption{Force ambiguity. Two groups of forces have the same resultant. }
    	\label{fig:force-ambi}
    \end{minipage}
\end{figure}

\paragraph{Tip force modeling}
The force $\boldsymbol{F}_{i}$, which is exerted by the tip $i$ on the object, is comprised of pressure and friction components which satisfy the Coulomb's Law of Friction.
All possible $\boldsymbol{F}_{i}$ form a friction cone at the contact point $\boldsymbol p_i$ (see Fig.~\ref{fig:fri-cone}).
For the tip that is not in contact with the object, we constrain the force in the friction cone centered at the tip-to-object projection point $\boldsymbol p_i$ (though the magnitude of the force can be nearly zero).
%
For computational convenience, we use a polyhedral cone as a linearized approximation of the fiction cone. 
Then, $\boldsymbol{F}_{i}$ can be expressed by the positive span of four normalized forces at the edges, which we denote as $\boldsymbol x_1,\cdots,\boldsymbol x_4$ (see Fig.~\ref{fig:fri-cone}). 
Let basis matrix $\boldsymbol{A}_i=[\boldsymbol{x}_1, \boldsymbol{x}_2, \boldsymbol{x}_3, \boldsymbol{x}_4]$, then the tip force can be computed as $\boldsymbol{F}_{i} = \boldsymbol{A}_i\boldsymbol{f}_i\ (\boldsymbol{f}_i \in \mathbb R^4, \boldsymbol{f}_i \ge 0)$. 
To this end, we use $\boldsymbol f_i$ as the optimization variable for the tip forces in Eq.~\ref{mainEnergyOfStageII}, which simplifies the friction cone constraints.
The friction coefficient is empirically set to $\mu=0.7$.
\paragraph{Force and moment term}
$E_{\text{f}}(\boldsymbol{f}_i)$ in Eq.~\ref{mainEnergyOfStageII} requires the object's linear acceleration to be explained by the resultant force including the tip forces and the gravity:
\begin{equation}
E_{\text{f}}(\boldsymbol{f}_i)=||\sum_{i=0}^4 \boldsymbol{A}_i\boldsymbol{f}_i+m_o\boldsymbol{g}-m_o\dot{\boldsymbol{v}}^t||^2,
\end{equation}
where $\boldsymbol{g}$ is the gravitational acceleration. Similarly,
$E_{\text{m}}(\boldsymbol{f}_i)$ in Eq.~\ref{mainEnergyOfStageII} requires the object's angular acceleration to be explained by the resultant moment:
\begin{equation}
E_{\text{m}}(\boldsymbol{f}_i) = ||\sum_{i=0}^4((\boldsymbol{p}_i-\boldsymbol{c}_o^t)\times (\boldsymbol{A}_i\boldsymbol{f}_i))-\boldsymbol{\tau}^t||^2,
\end{equation}
where $\boldsymbol{c}_o^t=\boldsymbol{W}^t\boldsymbol{c}_o$ is the center of mass of the live object, $\boldsymbol{\tau}^t$ is the driving torque of the object at frame $t$, which is derived from:
\begin{equation}
    \boldsymbol{\tau}^t = \boldsymbol{I}\dot{\boldsymbol{\omega}}^t + [\boldsymbol{\omega}^t] \boldsymbol{I} \boldsymbol{\omega}^t,
\end{equation}
where $[\cdot]$ means the cross product matrix.

\paragraph{Regularization term}
We introduce a regularization term $E_{\text{reg}}(\boldsymbol{f}_i)$ in Eq.~\ref{mainEnergyOfStageII} to confine the magnitude of the tip forces to be small.
This is based on two observations: \textit{1)} under specific situations, for example, when two contact points are on the opposite sides and forces are collinear, there are infinite numbers of solutions with arbitrarily large force values. Thus, we need a regularization term to ensure stable force solutions. \textit{2)} Humans are used to manipulating objects with the least possible forces.
The regularization term is defined as:
\begin{equation}
    E_{\text{reg}}(\boldsymbol{f}_i) = \sum_{i=0}^4||\boldsymbol{f}_i||^2.
\end{equation}

\paragraph{Contact term}
With the aforementioned terms, a group of forces that best explains the object's movement can be obtained. However, those terms do not model the contacts, \textit{i.e.,} every tip can exert any large forces on the object even if not in contact. Therefore, we introduce the contact term $E_{\text{tac}}(\boldsymbol{f}_i,\tilde{d}_i)$ in Eq.~\ref{mainEnergyOfStageII}:
\begin{equation}
    E_{\text{tac}}(\boldsymbol{f}_i,\tilde{d}_i) = \sum_{i=0}^4||\tilde{d}_i \boldsymbol{f}_i||^2,
\end{equation}
which means a tip can only exert little force ($\boldsymbol{f}_i$ is small) on the object when it is far from the object surface ($\tilde{d}_i$ is large), and vise versa.
This design is simple but important to our system.
Originally, updating the contact requires solving a discontinuous optimization at the boundary of contact and not contact.
With this design, we formulate it in a continuous optimization without involving noticeable errors.
To further constrain the solution, we add another energy term $E_{\text{smo}}(\tilde{d}_i)$ in Eq.~\ref{mainEnergyOfStageII} to penalize the difference between the optimized contact status and the initial contact status extracted from the kinematic tracking:
\begin{equation}
    E_{\text{smo}}(\tilde{d}_i) = \sum_{i=0}^4(\tilde{d}_i - d_i)^2.
\end{equation}

\subsection{Contact Movement Modeling}\label{sec:stage3}
With the refined contact status and force estimation, we model the hand-object movement at the contact points in this stage.
Since all interaction is done through contacts, correctly modeling the contact movement not only ensures the physical plausibility of the reconstruction, but also provides movement priors between the object and the hand, which greatly helps us to resolve the ambiguity in the insufficient input data caused by occlusions.
Nevertheless, the hand-object movement at the contacts can be very complex: the object can be static, sliding, or rolling at a contact; a soft body can deform at a contact.
In our method, we simplify the cases and focus on modeling the most common type of contact movement, \textit{i.e.,} the static and sliding contact.
\par
In the kinematic tracking result, the tip often slides on the object in fast motion when serious blurs exist in the input depth data. 
Simply forbidding any slip at the contact point will not give a satisfying result.
This is because human hands are flexible and can often slide on an object surface when manipulating it.
To address this, we propose a physics-based slide prevention method, which utilizes the force predicted in Stage II to constrain the slippage of the contact point with large pressure.
%
This means that the sliding contact is allowed when the pressure is small, which significantly improves the quality and physical plausibility of the interaction.
Moreover, the force estimation may be incorrect due to the ambiguities such as multiple contact points as shown in Fig. \ref{fig:force-ambi}.
On the other hand, kinematic tracking may be correct when there is sufficient observation.
Considering this, we additionally introduce the tip confidence extracted from the kinematic tracking.
In the following, we introduce the confidence extraction and the slide prevention respectively.

\subsubsection{Kinematic Confidence Extraction}
For each tip, a confidence value $C_i$ is computed from the depth map $\mathcal D^t$ and the kinematic tracking results $\boldsymbol \theta_{\mathrm{kin}}^t$ based on the number of observed depth points. To be specific, we predefine a number $N_s=75$ standing for the number of observed depth points that is sufficient for pose estimation of a tip by the kinematic method. Then we check the number of depth points $N_i$ that are close enough (distance smaller than 3mm) to the tip $i$. Then $C_i$ is defined as:
\begin{align}
     C_i = & \min(1, N_i/N_s).
\end{align}
Higher confidence indicates that the kinematic tracking gets sufficient data to fit the tip, which yields accurate results. On the contrary, lower confidence means the tip is partly or totally invisible. Leveraging the extracted confidence, our system reduces errors in the physics-based slide prevention, as detailed in the following.

\subsubsection{Confidence-based Slide Prevention}
The inputs of this module are the five tip positions after the contact status refinement $\boldsymbol{T}^{(\text{r})t}_i=\boldsymbol{p}_i + \tilde{d}_i \boldsymbol{n}_i$, the predicted pressure on each contact point $F_{N_i}=\boldsymbol{n}_i^T \boldsymbol{A}_i \boldsymbol{f}_i$, and the tip confidence $C_i$. 
The outputs of this module are the refined tip positions $\boldsymbol{T}^{(\text{s})t}_i$, which are used in a final inverse kinematics step.
We first compute the no-sliding tip positions $\boldsymbol{T}_i^{\mathrm{PS}}$ based on the final tip positions of the previous frame $\boldsymbol{T}^{(\text{f})t-1}_i$ by:
\begin{equation}
    \boldsymbol{T}^{\text{PS}}_i = \boldsymbol{T}^{(\text{r})t}_i - (\boldsymbol{I}_{3\times3} - \boldsymbol{n}_i \boldsymbol{n}_i^T)(\boldsymbol{T}^{(\text{r})t}_i - \boldsymbol W^t  (\boldsymbol W^{t-1})^{-1} \boldsymbol{T}^{(\text{f})t-1}_i),
\end{equation}
where $\boldsymbol{I}_{3\times3}$ is an identity matrix, $\boldsymbol{I}_{3\times3} - \boldsymbol{n}_i \boldsymbol{n}_i^T$ stands for a tangential projection on a surface point with the normal $\boldsymbol{n}_i$, $\boldsymbol W^t$ and $\boldsymbol W^{t-1}$ are the object poses at the current and the last frame, respectively.
In this way, the obtained $\boldsymbol{T}^{\text{PS}}_i$ has the same contact point as the previous frame but possible rigid and nonrigid deformation is allowed for the object. 
Then, we compute the refined tip positions $\boldsymbol{T}^{(\text{s})t}_i$ as:
\begin{equation}\label{eq:slide}
    \boldsymbol{T}^{(\text{s})t}_i = 
    \begin{cases}
    \boldsymbol{T}^{(\text{r})t}_i \quad& F_{N_i}<\alpha G_o \\ 
    \boldsymbol{T}^{\text{PS}}_i \quad& F_{N_i} \ge \alpha G_o \text{ and } \\ &C_i \le \frac{\beta}{||\boldsymbol{T}^{(\text{r})t}_i-\boldsymbol{T}^{\text{PS}}_i||} \\ 
    \gamma \boldsymbol{T}^{(\text{r})t}_i + (1-\gamma)\boldsymbol{T}^{\text{PS}}_i \quad& \text{otherwise}
    \end{cases},
\end{equation}
where $G_o=m_o||\boldsymbol{g}||$ is the object's gravity, $\alpha$ is a fixed scale parameter, $\beta$ is used to control the effect of the confidence $C_i$, and $\gamma$ is used to smooth the sliding process. 
We empirically set $\alpha = 0.3, \beta = 5\mathrm{mm}, \gamma = 0.5$.
We use a mass-dependent threshold $\alpha G_o$ to determine whether a tip can slide since the pressure $F_{N_i}$ solved in Stage II is proportional to $m_o$.
When sliding is allowed physically (the pressure is small, \textit{i.e.,} $F_{N_i} < \alpha G_o$ in Eq.~\ref{eq:slide}), we merely output the result $\boldsymbol{T}^{(\text{r})t}_i$ from the previous stage.
Otherwise, the tip movement along the contact tangential is restricted based on the kinematic confidence of the tip $C_i$. 
With low confidence, we tend to believe that the slide is caused by the tracking error, hence choose to prevent slippage (the middle row in Eq.~\ref{eq:slide}).
Otherwise, we choose to slide and perform a linear interpolation 
to smooth this process (the last row in Eq.~\ref{eq:slide}).
We use a dynamic threshold for confidence-based slide judgement. This is because even under fully observation ($C_i=1$), slippage is still inevitable in the kinematic result but only with small errors.
\par
Finally, we perform inverse kinematics to obtain the final hand pose $\boldsymbol{\theta}^t$ using the refined tip positions $\boldsymbol{T}^{(\text{s})t}_i$.
We also calculate the final tip positions $\boldsymbol{T}^{(\text{f})t}_i$ from $\boldsymbol{\theta}^t$, which are used in the next slide prevention stage of the next frame.
\section{Experiments}
In this section, we first provide the experimental settings of our system. Then, we compare our method with the state-of-the-art work \cite{zhang2021single} and evaluate the effectiveness of our key techniques qualitatively and quantitatively. Finally, we discuss our limitations. More results of rigid and nonrigid object motions of various object shapes, sizes, and textures are shown in Fig. \ref{fig:teaser} and the supplementary video.

\subsection{Experimental Settings}
We use one RealSense SR300 sensor to record the depth stream of hand-object interactions with the resolution of $320\times240$. Two NVIDIA TITANXp GPUs are used for the network evaluation and energy optimization respectively in the kinematic tracking stage. The physical refinement stage purely runs on an AMD Ryzen 5700g CPU. The contact status optimization Eq.~\ref{mainEnergyOfStageII} is solved by the LM algorithm \cite{more1978levenberg}. Our system can process one frame within 40ms, where 32ms for kinematic tracking and 8ms for physical refinement. 

\begin{figure}[!t]
	\centering
	\includegraphics[width=0.45\textwidth]{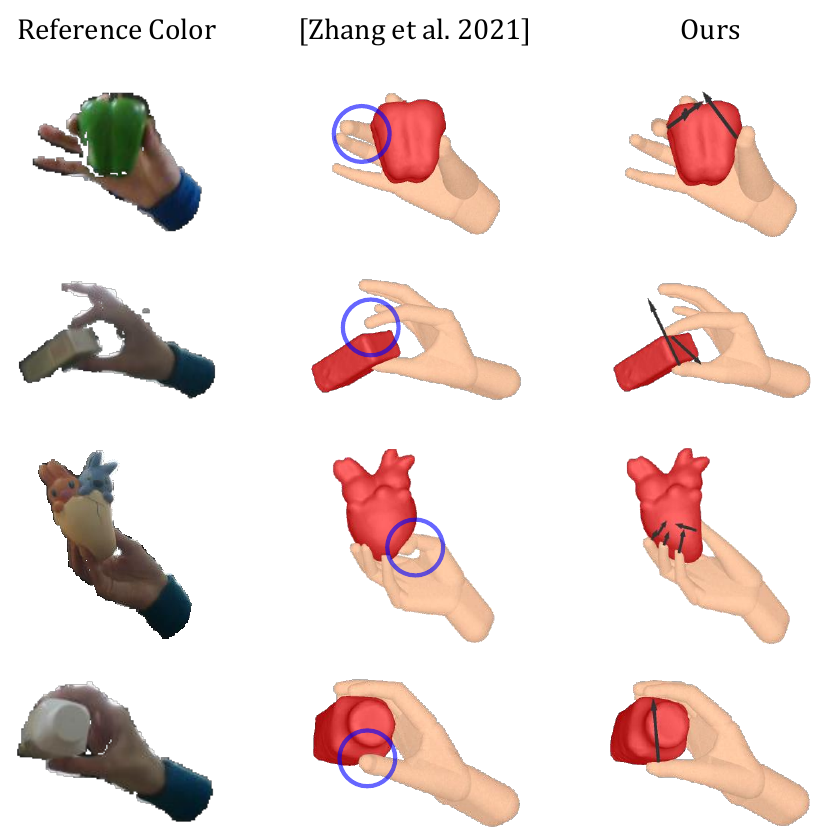}
	\caption{Qualitative comparison with \cite{zhang2021single}.}
	\label{fig:cmpWithZhang2021}
\end{figure}

\begin{figure}[!t]
\centering
    \includegraphics[width=0.45\textwidth]{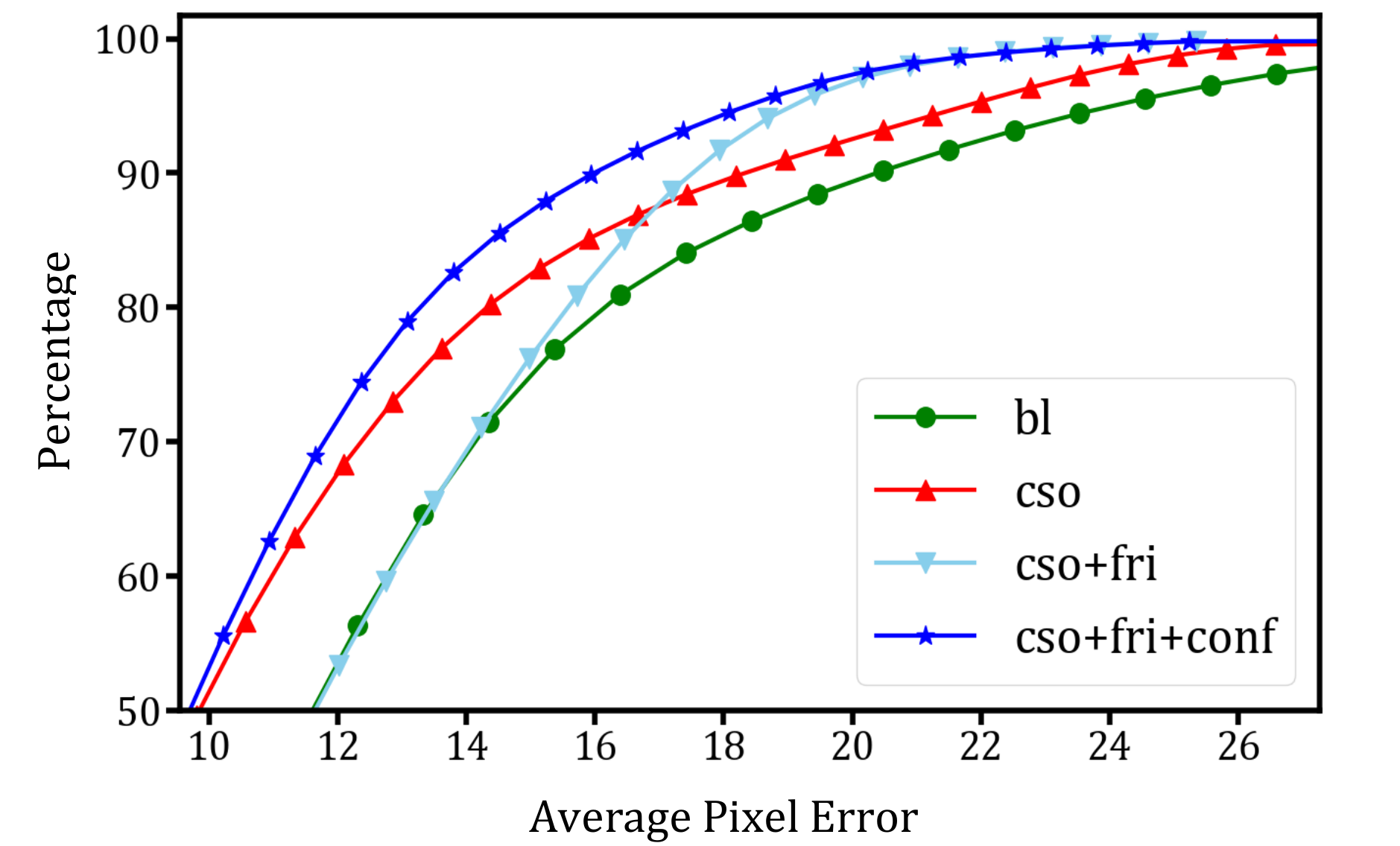}
    \caption{Percentage of correct hand tips. It stands for better performance if the curve is closer to the top-left corner.}
    \label{fig:quan-aggr-err-all}
\end{figure}

\begin{table}[!t]
\caption{Quantitative evaluation of different solutions using average pixel errors of all five fingertips in all frames.}
\label{tab:abl}
\begin{tabular}{ccccc}
\toprule
 &bl    & cso           & cso+fri & cso+fri+conf   \\
\midrule
Average Pixel Error &11.58 & 10.22 & 11.20    & \textbf{9.67}           \\
\bottomrule
\end{tabular}
\end{table}

\begin{table}
\caption{Proportion of physically implausible frames (less than 2 contacts).}
\label{tab:cmp-phys-plau}
\begin{tabular}{ccc}
\toprule
 &\cite{zhang2021single}    & Ours   \\
\midrule
Implausible Ratio (\%) &54.7 & \textbf{2.0}           \\
\bottomrule
\end{tabular}
\end{table}

\subsection{Comparisons}
Our work uses \cite{zhang2021single} for kinematic tracking which is also the state-of-the-art in this topic, so we compare our system with it to demonstrate the effectiveness of our physics-based refinement on precision and physical plausibility.
The qualitative results are shown in the supplementary video and Fig.~\ref{fig:cmpWithZhang2021}. 
In the top two rows of Fig.~\ref{fig:cmpWithZhang2021}, we can clearly see that sometimes a contact may not be discovered by \cite{zhang2021single} while we correctly reconstruct it by the physics-based contact status optimization. 
Meanwhile, in the bottom two rows of Fig.~\ref{fig:cmpWithZhang2021}, some contacts may slide to a wrong position in the result of \cite{zhang2021single} due to the lack of observation. 
While in our result, as we model the friction, the contact position is better preserved. 
Note that our method can prevent error accumulation in the kinematic tracking system (\textit{e.g.}, the third row in Fig.~\ref{fig:cmpWithZhang2021}). This is because the hand pose solved in the previous frame is used to initialize the next frame. Thus, the physical refinement can influence the solution in the subsequent frames.
More visual comparisons can be found in the supplementary video.

For a quantitative comparison on precision, we use the same metric as \cite{zhang2021single}. 
We project five fingertips of the reconstruction result onto the RGB camera and manually label the tips on the color images as ground truth. 
The average pixel error is used to measure the accuracy of hand tracking. 
We prepare three annotated sequences for this comparison: "MoveBox" with 480 frames, "RotateBottle" with 390 frames, and "HandleToy" with 550 frames. 
Tab.~\ref{tab:abl} shows the values of average pixel errors on the three recorded sequences and Fig.~\ref{fig:quan-aggr-err-all} reports the corresponding aggregated errors.
\textit{bl} stands for our kinematic tracking result which is just the result of \cite{zhang2021single}. 
These comparisons further demonstrate that our method can improve the tracking accuracy of fingertips and lessen the maximum error as the curve in Fig.~\ref{fig:quan-aggr-err-all} is more on the left.  

To demonstrate that our results are more physically plausible, we compute the ratio of frames with less than two contact points in the aforementioned three sequences. As multiple contact points are necessary for supporting an object in most cases, we regard circumstances violating this as physically implausible.
Tab.~\ref{tab:cmp-phys-plau} shows that our method can significantly reduce the implausible cases by adding sufficient contact points based on our physical model.

\subsection{Evaluation}
We first quantitatively evaluate the effectiveness of the key components in our physical refinement, which is also shown in Tab.~\ref{tab:abl} and Fig.~\ref{fig:quan-aggr-err-all}. 
Here we show the results of four solutions, pure kinematic results (\textit{bl}), contact status optimization only (\textit{cso}), contact status optimization + pure friction-based slide prevention (\textit{cso+fri}), and contact status optimization + confidence-based slide prevention (\textit{cso+fri+conf}). 
Contact status optimization (\textit{cso}) helps to resolve the problem that fingertips fail to properly contact the object and it reduces the numerical errors noticeably.
However, adding the friction-based sliding prevention (\textit{cso+fri}) cannot further reduce the errors but increase them a little bit.
This is because this module is purely based on physics without considering the kinematic information.
If the initial contact position has errors, fixing its position by friction makes the contact position impossible to be refined, leading to worse results sometimes. 
As in the case with strong observations, kinematic results are usually correct.
So, after adding the kinematic-involved confidence component (\textit{cso+fri+conf}), the errors dropped and the best performance in this experiment is achieved.
\par
\begin{figure}[!t]
\centering
    \includegraphics[width=0.45\textwidth]{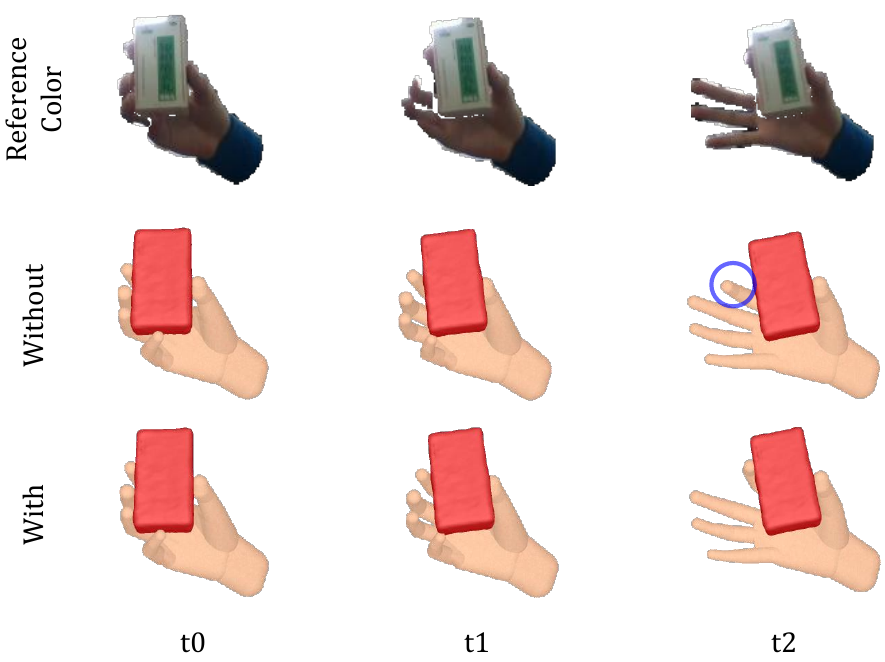}
    \caption{Qualitative evaluation of our contact status optimization.}
    \label{fig:quali-abl-cso}
\end{figure}

We also show some qualitative results here.
In Fig.~\ref{fig:quali-abl-cso}, we see that the kinematic reconstruction fails to reconstruct the left contact point due to the relatively fast motion. The index finger wrongly fits the observation of the middle finger and moves wrongly with the middle finger. 
On the other hand, \textit{cso} knows that one contact cannot prevent the object from falling down and correctly discovers the second contact point.
More results are shown in the supplementary video.
Fig.~\ref{fig:quali-abl-fri} gives one visual example to show the effectiveness of our confidence-based slide prevention. In these snapshots of a short sequence, we see that without this slide prevention, the thumb wrongly slides on the object surface due to the lack of observations when the object turns at frames around t1 and t2. 
Notice that the involvement of kinematic confidence not only fixes the contact when pressure exists, but also allows sliding when no pressure exists. To better demonstrate this, we show the result in the supplementary video in the section "Evaluation of the Confidence".

\begin{figure*}[!t]
\centering
    \includegraphics[width=0.9\textwidth]{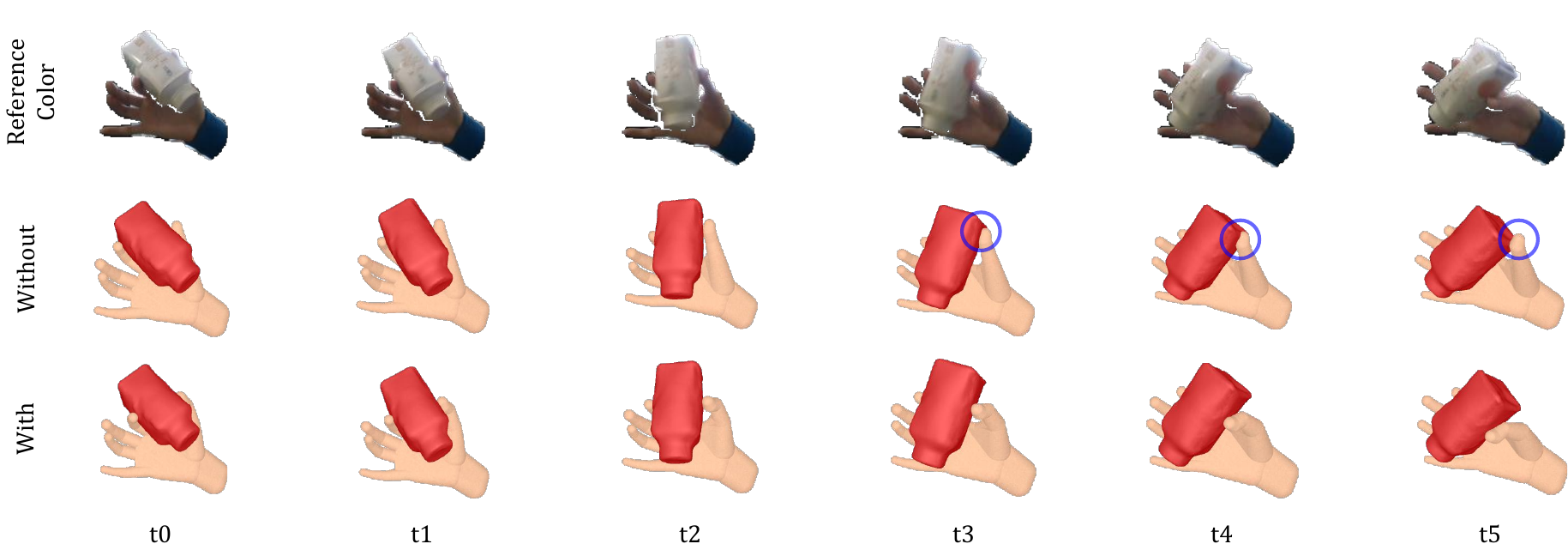}
    \caption{Qualitative evaluation of our slide prevention in a sequence.}
    \label{fig:quali-abl-fri}
\end{figure*}

\subsection{Limitations}
To guarantee real-time performance, our physical model is largely simplified and thus still far from real physics. 
First, we assume the contacts only happens on fingertips. For cases breaking this assumption, our system goes back to a kinematic tracking system. 
Second, we solve for the contact forces with assumptions on object masses. 
This may lead to inaccurate contact force magnitudes, but still successfully brings physical plausibility into the reconstruction system.
Third, when tracking non-rigid objects, we fix inertia information instead of updating it with the object's deformation. In most cases, this only affects the accuracy of the force solution. However, for strong deformations, this simplification may lead to wrong contact positions due to the big errors in the physical properties.




\section{Conclusions}
This paper proposes a physical method for modeling the complex motion of hand-object interaction and demonstrates its power in single view real-time reconstruction. 
The method first models the relationship between the contact forces and the dynamics of the in-hand object, which is successfully used in recovering unobserved contacts as well as the contact forces. 
It also jointly models the static and sliding contact motions, which are used to reconstruct accurate contact motions by combining both the kinematic confidence and the contact forces.
In general, this technique achieves physically plausible and more accurate interaction reconstruction in real-time using a single view input, with a byproduct of plausible forces.


\begin{acks}
This work was supported by Beijing Natural Science Foundation (JQ19015), the NSFC (No.62021002, 61727808), the National Key R\&D Program of China 2018YFA0704000, the Key Research and Development Project of Tibet Autonomous Region XZ202101ZY0019G. This work was supported by THUIBCS, Tsinghua University and BLBCI, Beijing Municipal Education Commission. Feng Xu and Jun-Hai Yong are the corresponding authors. Hao Zhang contributed to this work when he was a PhD student in Tsinghua University.
\end{acks}

\bibliography{ref}


\begin{thebibliography}{41}


\ifx \showCODEN    \undefined \def \showCODEN     #1{\unskip}     \fi
\ifx \showDOI      \undefined \def \showDOI       #1{#1}\fi
\ifx \showISBNx    \undefined \def \showISBNx     #1{\unskip}     \fi
\ifx \showISBNxiii \undefined \def \showISBNxiii  #1{\unskip}     \fi
\ifx \showISSN     \undefined \def \showISSN      #1{\unskip}     \fi
\ifx \showLCCN     \undefined \def \showLCCN      #1{\unskip}     \fi
\ifx \shownote     \undefined \def \shownote      #1{#1}          \fi
\ifx \showarticletitle \undefined \def \showarticletitle #1{#1}   \fi
\ifx \showURL      \undefined \def \showURL       {\relax}        \fi
\providecommand\bibfield[2]{#2}
\providecommand\bibinfo[2]{#2}
\providecommand\natexlab[1]{#1}
\providecommand\showeprint[2][]{arXiv:#2}

\bibitem[\protect\citeauthoryear{Ballan, Taneja, Gall, Gool, and
  Pollefeys}{Ballan et~al\mbox{.}}{2012}]%
        {ballan2012motion}
\bibfield{author}{\bibinfo{person}{Luca Ballan}, \bibinfo{person}{Aparna
  Taneja}, \bibinfo{person}{J{\"u}rgen Gall}, \bibinfo{person}{Luc~Van Gool},
  {and} \bibinfo{person}{Marc Pollefeys}.} \bibinfo{year}{2012}\natexlab{}.
\newblock \showarticletitle{Motion capture of hands in action using
  discriminative salient points}. In \bibinfo{booktitle}{\emph{European
  Conference on Computer Vision}}. Springer, \bibinfo{pages}{640--653}.
\newblock


\bibitem[\protect\citeauthoryear{Christen, Kocabas, Aksan, Hwangbo, Song, and
  Hilliges}{Christen et~al\mbox{.}}{2022}]%
        {christen2022d}
\bibfield{author}{\bibinfo{person}{Sammy Christen}, \bibinfo{person}{Muhammed
  Kocabas}, \bibinfo{person}{Emre Aksan}, \bibinfo{person}{Jemin Hwangbo},
  \bibinfo{person}{Jie Song}, {and} \bibinfo{person}{Otmar Hilliges}.}
  \bibinfo{year}{2022}\natexlab{}.
\newblock \showarticletitle{D-Grasp: Physically Plausible Dynamic Grasp
  Synthesis for Hand-Object Interactions}. In
  \bibinfo{booktitle}{\emph{Proceedings of the IEEE/CVF Conference on Computer
  Vision and Pattern Recognition}}. \bibinfo{pages}{20577--20586}.
\newblock


\bibitem[\protect\citeauthoryear{Ehsani, Tulsiani, Gupta, Farhadi, and
  Gupta}{Ehsani et~al\mbox{.}}{2020}]%
        {ehsani2020use}
\bibfield{author}{\bibinfo{person}{Kiana Ehsani}, \bibinfo{person}{Shubham
  Tulsiani}, \bibinfo{person}{Saurabh Gupta}, \bibinfo{person}{Ali Farhadi},
  {and} \bibinfo{person}{Abhinav Gupta}.} \bibinfo{year}{2020}\natexlab{}.
\newblock \showarticletitle{Use the force, luke! learning to predict physical
  forces by simulating effects}. In \bibinfo{booktitle}{\emph{Proceedings of
  the IEEE/CVF Conference on Computer Vision and Pattern Recognition}}.
  \bibinfo{pages}{224--233}.
\newblock


\bibitem[\protect\citeauthoryear{Grady, Tang, Twigg, Vo, Brahmbhatt, and
  Kemp}{Grady et~al\mbox{.}}{2021}]%
        {grady2021contactopt}
\bibfield{author}{\bibinfo{person}{Patrick Grady}, \bibinfo{person}{Chengcheng
  Tang}, \bibinfo{person}{Christopher~D Twigg}, \bibinfo{person}{Minh Vo},
  \bibinfo{person}{Samarth Brahmbhatt}, {and} \bibinfo{person}{Charles~C
  Kemp}.} \bibinfo{year}{2021}\natexlab{}.
\newblock \showarticletitle{Contactopt: Optimizing contact to improve grasps}.
  In \bibinfo{booktitle}{\emph{Proceedings of the IEEE/CVF Conference on
  Computer Vision and Pattern Recognition}}. \bibinfo{pages}{1471--1481}.
\newblock


\bibitem[\protect\citeauthoryear{Hasson, Varol, Tzionas, Kalevatykh, Black,
  Laptev, and Schmid}{Hasson et~al\mbox{.}}{2019}]%
        {hasson2019learning}
\bibfield{author}{\bibinfo{person}{Yana Hasson}, \bibinfo{person}{Gul Varol},
  \bibinfo{person}{Dimitrios Tzionas}, \bibinfo{person}{Igor Kalevatykh},
  \bibinfo{person}{Michael~J Black}, \bibinfo{person}{Ivan Laptev}, {and}
  \bibinfo{person}{Cordelia Schmid}.} \bibinfo{year}{2019}\natexlab{}.
\newblock \showarticletitle{Learning joint reconstruction of hands and
  manipulated objects}. In \bibinfo{booktitle}{\emph{Proceedings of the
  IEEE/CVF conference on computer vision and pattern recognition}}.
  \bibinfo{pages}{11807--11816}.
\newblock


\bibitem[\protect\citeauthoryear{Hirota and Tagawa}{Hirota and Tagawa}{2016}]%
        {hirota2016interaction}
\bibfield{author}{\bibinfo{person}{Koichi Hirota} {and}
  \bibinfo{person}{Kazuyoshi Tagawa}.} \bibinfo{year}{2016}\natexlab{}.
\newblock \showarticletitle{Interaction with virtual object using deformable
  hand}. In \bibinfo{booktitle}{\emph{2016 IEEE Virtual Reality (VR)}}. IEEE,
  \bibinfo{pages}{49--56}.
\newblock


\bibitem[\protect\citeauthoryear{H{\"o}ll, Oberweger, Arth, and
  Lepetit}{H{\"o}ll et~al\mbox{.}}{2018}]%
        {holl2018efficient}
\bibfield{author}{\bibinfo{person}{Markus H{\"o}ll}, \bibinfo{person}{Markus
  Oberweger}, \bibinfo{person}{Clemens Arth}, {and} \bibinfo{person}{Vincent
  Lepetit}.} \bibinfo{year}{2018}\natexlab{}.
\newblock \showarticletitle{Efficient physics-based implementation for
  realistic hand-object interaction in virtual reality}. In
  \bibinfo{booktitle}{\emph{2018 IEEE Conference on Virtual Reality and 3D User
  Interfaces (VR)}}. IEEE, \bibinfo{pages}{175--182}.
\newblock


\bibitem[\protect\citeauthoryear{Isogawa, Yuan, O'Toole, and Kitani}{Isogawa
  et~al\mbox{.}}{2020}]%
        {Isogawa2020}
\bibfield{author}{\bibinfo{person}{Mariko Isogawa}, \bibinfo{person}{Ye Yuan},
  \bibinfo{person}{Matthew O'Toole}, {and} \bibinfo{person}{Kris Kitani}.}
  \bibinfo{year}{2020}\natexlab{}.
\newblock \showarticletitle{Optical Non-Line-of-Sight Physics-Based 3D Human
  Pose Estimation}. In \bibinfo{booktitle}{\emph{2020 IEEE/CVF Conference on
  Computer Vision and Pattern Recognition (CVPR)}}.
\newblock


\bibitem[\protect\citeauthoryear{Kry and Pai}{Kry and Pai}{2006}]%
        {kry2006interaction}
\bibfield{author}{\bibinfo{person}{Paul~G Kry} {and} \bibinfo{person}{Dinesh~K
  Pai}.} \bibinfo{year}{2006}\natexlab{}.
\newblock \showarticletitle{Interaction capture and synthesis}.
\newblock \bibinfo{journal}{\emph{ACM Transactions on Graphics (TOG)}}
  \bibinfo{volume}{25}, \bibinfo{number}{3} (\bibinfo{year}{2006}),
  \bibinfo{pages}{872--880}.
\newblock


\bibitem[\protect\citeauthoryear{Kumar, Vaidya, and Huth}{Kumar
  et~al\mbox{.}}{2021}]%
        {kumar2021physically}
\bibfield{author}{\bibinfo{person}{Akarsh Kumar}, \bibinfo{person}{Aditya~R
  Vaidya}, {and} \bibinfo{person}{Alexander~G Huth}.}
  \bibinfo{year}{2021}\natexlab{}.
\newblock \showarticletitle{Physically Plausible Pose Refinement using Fully
  Differentiable Forces}.
\newblock \bibinfo{journal}{\emph{arXiv preprint arXiv:2105.08196}}
  (\bibinfo{year}{2021}).
\newblock


\bibitem[\protect\citeauthoryear{Kyriazis and Argyros}{Kyriazis and
  Argyros}{2014}]%
        {kyriazis2014scalable}
\bibfield{author}{\bibinfo{person}{Nikolaos Kyriazis} {and}
  \bibinfo{person}{Antonis Argyros}.} \bibinfo{year}{2014}\natexlab{}.
\newblock \showarticletitle{Scalable 3d tracking of multiple interacting
  objects}. In \bibinfo{booktitle}{\emph{Proceedings of the IEEE Conference on
  Computer Vision and Pattern Recognition}}. \bibinfo{pages}{3430--3437}.
\newblock


\bibitem[\protect\citeauthoryear{Li, Sedlar, Carpentier, Laptev, Mansard, and
  Sivic}{Li et~al\mbox{.}}{2019}]%
        {Li2019}
\bibfield{author}{\bibinfo{person}{Zongmian Li}, \bibinfo{person}{Jiri Sedlar},
  \bibinfo{person}{Justin Carpentier}, \bibinfo{person}{Ivan Laptev},
  \bibinfo{person}{Nicolas Mansard}, {and} \bibinfo{person}{Josef Sivic}.}
  \bibinfo{year}{2019}\natexlab{}.
\newblock \showarticletitle{Estimating 3D Motion and Forces of Person-Object
  Interactions from Monocular Video}. In \bibinfo{booktitle}{\emph{Computer
  Vision and Pattern Recognition (CVPR)}}.
\newblock


\bibitem[\protect\citeauthoryear{Liu}{Liu}{2009}]%
        {liu2009dextrous}
\bibfield{author}{\bibinfo{person}{C~Karen Liu}.}
  \bibinfo{year}{2009}\natexlab{}.
\newblock \showarticletitle{Dextrous manipulation from a grasping pose}.
\newblock In \bibinfo{booktitle}{\emph{ACM SIGGRAPH 2009 papers}}.
  \bibinfo{pages}{1--6}.
\newblock


\bibitem[\protect\citeauthoryear{Mor{\'e}}{Mor{\'e}}{1978}]%
        {more1978levenberg}
\bibfield{author}{\bibinfo{person}{Jorge~J Mor{\'e}}.}
  \bibinfo{year}{1978}\natexlab{}.
\newblock \showarticletitle{The Levenberg-Marquardt algorithm: implementation
  and theory}.
\newblock In \bibinfo{booktitle}{\emph{Numerical analysis}}.
  \bibinfo{publisher}{Springer}, \bibinfo{pages}{105--116}.
\newblock


\bibitem[\protect\citeauthoryear{Newcombe, Fox, and Seitz}{Newcombe
  et~al\mbox{.}}{2015}]%
        {newcombe2015dynamicfusion}
\bibfield{author}{\bibinfo{person}{Richard~A Newcombe}, \bibinfo{person}{Dieter
  Fox}, {and} \bibinfo{person}{Steven~M Seitz}.}
  \bibinfo{year}{2015}\natexlab{}.
\newblock \showarticletitle{Dynamicfusion: Reconstruction and tracking of
  non-rigid scenes in real-time}. In \bibinfo{booktitle}{\emph{Proceedings of
  the IEEE conference on computer vision and pattern recognition}}.
  \bibinfo{pages}{343--352}.
\newblock


\bibitem[\protect\citeauthoryear{Oikonomidis, Kyriazis, and
  Argyros}{Oikonomidis et~al\mbox{.}}{2011}]%
        {oikonomidis2011full}
\bibfield{author}{\bibinfo{person}{Iason Oikonomidis},
  \bibinfo{person}{Nikolaos Kyriazis}, {and} \bibinfo{person}{Antonis~A
  Argyros}.} \bibinfo{year}{2011}\natexlab{}.
\newblock \showarticletitle{Full dof tracking of a hand interacting with an
  object by modeling occlusions and physical constraints}. In
  \bibinfo{booktitle}{\emph{2011 International Conference on Computer Vision}}.
  IEEE, \bibinfo{pages}{2088--2095}.
\newblock


\bibitem[\protect\citeauthoryear{Panteleris and Argyros}{Panteleris and
  Argyros}{2017}]%
        {panteleris2017back}
\bibfield{author}{\bibinfo{person}{Paschalis Panteleris} {and}
  \bibinfo{person}{Antonis Argyros}.} \bibinfo{year}{2017}\natexlab{}.
\newblock \showarticletitle{Back to rgb: 3d tracking of hands and hand-object
  interactions based on short-baseline stereo}. In
  \bibinfo{booktitle}{\emph{Proceedings of the IEEE International Conference on
  Computer Vision Workshops}}. \bibinfo{pages}{575--584}.
\newblock


\bibitem[\protect\citeauthoryear{Panteleris, Kyriazis, and Argyros}{Panteleris
  et~al\mbox{.}}{2015}]%
        {BMVC2015_123}
\bibfield{author}{\bibinfo{person}{Paschalis Panteleris},
  \bibinfo{person}{Nikolaos Kyriazis}, {and} \bibinfo{person}{Antonis~A.
  Argyros}.} \bibinfo{year}{2015}\natexlab{}.
\newblock \showarticletitle{3D Tracking of Human Hands in Interaction with
  Unknown Objects}. In \bibinfo{booktitle}{\emph{Proceedings of the British
  Machine Vision Conference (BMVC)}}. Article \bibinfo{articleno}{123},
  \bibinfo{numpages}{12}~pages.
\newblock


\bibitem[\protect\citeauthoryear{Peng, Abbeel, Levine, and van~de Panne}{Peng
  et~al\mbox{.}}{2018a}]%
        {DeepMimic}
\bibfield{author}{\bibinfo{person}{Xue~Bin Peng}, \bibinfo{person}{Pieter
  Abbeel}, \bibinfo{person}{Sergey Levine}, {and} \bibinfo{person}{Michiel
  van~de Panne}.} \bibinfo{year}{2018}\natexlab{a}.
\newblock \showarticletitle{DeepMimic: Example-Guided Deep Reinforcement
  Learning of Physics-Based Character Skills}.
\newblock \bibinfo{journal}{\emph{ACM Trans. Graph.}}  \bibinfo{volume}{37}
  (\bibinfo{date}{jul} \bibinfo{year}{2018}).
\newblock


\bibitem[\protect\citeauthoryear{Peng, Kanazawa, Malik, Abbeel, and
  Levine}{Peng et~al\mbox{.}}{2018b}]%
        {SFV}
\bibfield{author}{\bibinfo{person}{Xue~Bin Peng}, \bibinfo{person}{Angjoo
  Kanazawa}, \bibinfo{person}{Jitendra Malik}, \bibinfo{person}{Pieter Abbeel},
  {and} \bibinfo{person}{Sergey Levine}.} \bibinfo{year}{2018}\natexlab{b}.
\newblock \showarticletitle{SFV: Reinforcement Learning of Physical Skills from
  Videos}.
\newblock \bibinfo{journal}{\emph{ACM Trans. Graph.}}  \bibinfo{volume}{37}
  (\bibinfo{date}{nov} \bibinfo{year}{2018}).
\newblock


\bibitem[\protect\citeauthoryear{Pham, Kheddar, Qammaz, and Argyros}{Pham
  et~al\mbox{.}}{2015}]%
        {pham2015towards}
\bibfield{author}{\bibinfo{person}{Tu-Hoa Pham}, \bibinfo{person}{Abderrahmane
  Kheddar}, \bibinfo{person}{Ammar Qammaz}, {and} \bibinfo{person}{Antonis~A
  Argyros}.} \bibinfo{year}{2015}\natexlab{}.
\newblock \showarticletitle{Towards force sensing from vision: Observing
  hand-object interactions to infer manipulation forces}. In
  \bibinfo{booktitle}{\emph{Proceedings of the IEEE conference on computer
  vision and pattern recognition}}. \bibinfo{pages}{2810--2819}.
\newblock


\bibitem[\protect\citeauthoryear{Pham, Kyriazis, Argyros, and Kheddar}{Pham
  et~al\mbox{.}}{2017}]%
        {pham2017hand}
\bibfield{author}{\bibinfo{person}{Tu-Hoa Pham}, \bibinfo{person}{Nikolaos
  Kyriazis}, \bibinfo{person}{Antonis~A Argyros}, {and}
  \bibinfo{person}{Abderrahmane Kheddar}.} \bibinfo{year}{2017}\natexlab{}.
\newblock \showarticletitle{Hand-object contact force estimation from
  markerless visual tracking}.
\newblock \bibinfo{journal}{\emph{IEEE transactions on pattern analysis and
  machine intelligence}} \bibinfo{volume}{40}, \bibinfo{number}{12}
  (\bibinfo{year}{2017}), \bibinfo{pages}{2883--2896}.
\newblock


\bibitem[\protect\citeauthoryear{Pollard and Zordan}{Pollard and
  Zordan}{2005}]%
        {pollard2005physically}
\bibfield{author}{\bibinfo{person}{Nancy~S Pollard} {and}
  \bibinfo{person}{Victor~Brian Zordan}.} \bibinfo{year}{2005}\natexlab{}.
\newblock \showarticletitle{Physically based grasping control from example}. In
  \bibinfo{booktitle}{\emph{Proceedings of the 2005 ACM SIGGRAPH/Eurographics
  symposium on Computer animation}}. \bibinfo{pages}{311--318}.
\newblock


\bibitem[\protect\citeauthoryear{Rempe, Guibas, Hertzmann, Russell, Villegas,
  and Yang}{Rempe et~al\mbox{.}}{2020}]%
        {Rempe2020}
\bibfield{author}{\bibinfo{person}{Davis Rempe}, \bibinfo{person}{Leonidas~J.
  Guibas}, \bibinfo{person}{Aaron Hertzmann}, \bibinfo{person}{Bryan Russell},
  \bibinfo{person}{Ruben Villegas}, {and} \bibinfo{person}{Jimei Yang}.}
  \bibinfo{year}{2020}\natexlab{}.
\newblock \showarticletitle{Contact and Human Dynamics from Monocular Video}.
  In \bibinfo{booktitle}{\emph{Proceedings of the European Conference on
  Computer Vision (ECCV)}}.
\newblock


\bibitem[\protect\citeauthoryear{Shimada, Golyanik, Xu, P\'{e}rez, and
  Theobalt}{Shimada et~al\mbox{.}}{2021}]%
        {PhysAware}
\bibfield{author}{\bibinfo{person}{Soshi Shimada}, \bibinfo{person}{Vladislav
  Golyanik}, \bibinfo{person}{Weipeng Xu}, \bibinfo{person}{Patrick P\'{e}rez},
  {and} \bibinfo{person}{Christian Theobalt}.} \bibinfo{year}{2021}\natexlab{}.
\newblock \showarticletitle{Neural Monocular 3D Human Motion Capture with
  Physical Awareness}.
\newblock \bibinfo{journal}{\emph{ACM Transactions on Graphics}}
  \bibinfo{volume}{40} (\bibinfo{date}{aug} \bibinfo{year}{2021}).
\newblock


\bibitem[\protect\citeauthoryear{Shimada, Golyanik, Xu, and Theobalt}{Shimada
  et~al\mbox{.}}{2020}]%
        {PhysCap}
\bibfield{author}{\bibinfo{person}{Soshi Shimada}, \bibinfo{person}{Vladislav
  Golyanik}, \bibinfo{person}{Weipeng Xu}, {and} \bibinfo{person}{Christian
  Theobalt}.} \bibinfo{year}{2020}\natexlab{}.
\newblock \showarticletitle{PhysCap: physically plausible monocular 3D motion
  capture in real time}.
\newblock \bibinfo{journal}{\emph{ACM Transactions on Graphics}}
  \bibinfo{volume}{39} (\bibinfo{date}{dec} \bibinfo{year}{2020}).
\newblock


\bibitem[\protect\citeauthoryear{Sridhar, Mueller, Zollh{\"o}fer, Casas,
  Oulasvirta, and Theobalt}{Sridhar et~al\mbox{.}}{2016}]%
        {sridhar2016real}
\bibfield{author}{\bibinfo{person}{Srinath Sridhar}, \bibinfo{person}{Franziska
  Mueller}, \bibinfo{person}{Michael Zollh{\"o}fer}, \bibinfo{person}{Dan
  Casas}, \bibinfo{person}{Antti Oulasvirta}, {and} \bibinfo{person}{Christian
  Theobalt}.} \bibinfo{year}{2016}\natexlab{}.
\newblock \showarticletitle{Real-time joint tracking of a hand manipulating an
  object from rgb-d input}. In \bibinfo{booktitle}{\emph{European Conference on
  Computer Vision}}. Springer, \bibinfo{pages}{294--310}.
\newblock


\bibitem[\protect\citeauthoryear{Talvas, Marchal, Duriez, and Otaduy}{Talvas
  et~al\mbox{.}}{2015}]%
        {talvas2015aggregate}
\bibfield{author}{\bibinfo{person}{Anthony Talvas}, \bibinfo{person}{Maud
  Marchal}, \bibinfo{person}{Christian Duriez}, {and} \bibinfo{person}{Miguel~A
  Otaduy}.} \bibinfo{year}{2015}\natexlab{}.
\newblock \showarticletitle{Aggregate constraints for virtual manipulation with
  soft fingers}.
\newblock \bibinfo{journal}{\emph{IEEE transactions on visualization and
  computer graphics}} \bibinfo{volume}{21}, \bibinfo{number}{4}
  (\bibinfo{year}{2015}), \bibinfo{pages}{452--461}.
\newblock


\bibitem[\protect\citeauthoryear{Tekin, Bogo, and Pollefeys}{Tekin
  et~al\mbox{.}}{2019}]%
        {tekin2019h+}
\bibfield{author}{\bibinfo{person}{Bugra Tekin}, \bibinfo{person}{Federica
  Bogo}, {and} \bibinfo{person}{Marc Pollefeys}.}
  \bibinfo{year}{2019}\natexlab{}.
\newblock \showarticletitle{H+ o: Unified egocentric recognition of 3d
  hand-object poses and interactions}. In \bibinfo{booktitle}{\emph{Proceedings
  of the IEEE/CVF conference on computer vision and pattern recognition}}.
  \bibinfo{pages}{4511--4520}.
\newblock


\bibitem[\protect\citeauthoryear{Tkach, Pauly, and Tagliasacchi}{Tkach
  et~al\mbox{.}}{2016}]%
        {tkach2016sphere}
\bibfield{author}{\bibinfo{person}{Anastasia Tkach}, \bibinfo{person}{Mark
  Pauly}, {and} \bibinfo{person}{Andrea Tagliasacchi}.}
  \bibinfo{year}{2016}\natexlab{}.
\newblock \showarticletitle{Sphere-meshes for real-time hand modeling and
  tracking}.
\newblock \bibinfo{journal}{\emph{ACM Transactions on Graphics (ToG)}}
  \bibinfo{volume}{35}, \bibinfo{number}{6} (\bibinfo{year}{2016}),
  \bibinfo{pages}{1--11}.
\newblock


\bibitem[\protect\citeauthoryear{Tzionas, Ballan, Srikantha, Aponte, Pollefeys,
  and Gall}{Tzionas et~al\mbox{.}}{2016}]%
        {tzionas2016capturing}
\bibfield{author}{\bibinfo{person}{Dimitrios Tzionas}, \bibinfo{person}{Luca
  Ballan}, \bibinfo{person}{Abhilash Srikantha}, \bibinfo{person}{Pablo
  Aponte}, \bibinfo{person}{Marc Pollefeys}, {and} \bibinfo{person}{Juergen
  Gall}.} \bibinfo{year}{2016}\natexlab{}.
\newblock \showarticletitle{Capturing hands in action using discriminative
  salient points and physics simulation}.
\newblock \bibinfo{journal}{\emph{International Journal of Computer Vision}}
  \bibinfo{volume}{118}, \bibinfo{number}{2} (\bibinfo{year}{2016}),
  \bibinfo{pages}{172--193}.
\newblock


\bibitem[\protect\citeauthoryear{Wang, Min, Zhang, Liu, Xu, Dai, and Chai}{Wang
  et~al\mbox{.}}{2013}]%
        {wang2013video}
\bibfield{author}{\bibinfo{person}{Yangang Wang}, \bibinfo{person}{Jianyuan
  Min}, \bibinfo{person}{Jianjie Zhang}, \bibinfo{person}{Yebin Liu},
  \bibinfo{person}{Feng Xu}, \bibinfo{person}{Qionghai Dai}, {and}
  \bibinfo{person}{Jinxiang Chai}.} \bibinfo{year}{2013}\natexlab{}.
\newblock \showarticletitle{Video-based hand manipulation capture through
  composite motion control}.
\newblock \bibinfo{journal}{\emph{ACM Transactions on Graphics (TOG)}}
  \bibinfo{volume}{32}, \bibinfo{number}{4} (\bibinfo{year}{2013}),
  \bibinfo{pages}{1--14}.
\newblock


\bibitem[\protect\citeauthoryear{Yang, Yin, and Liu}{Yang
  et~al\mbox{.}}{2022}]%
        {yang2022learning}
\bibfield{author}{\bibinfo{person}{Zeshi Yang}, \bibinfo{person}{Kangkang Yin},
  {and} \bibinfo{person}{Libin Liu}.} \bibinfo{year}{2022}\natexlab{}.
\newblock \showarticletitle{Learning to use chopsticks in diverse gripping
  styles}.
\newblock \bibinfo{journal}{\emph{ACM Transactions on Graphics (TOG)}}
  \bibinfo{volume}{41}, \bibinfo{number}{4} (\bibinfo{year}{2022}),
  \bibinfo{pages}{1--17}.
\newblock


\bibitem[\protect\citeauthoryear{Yi, Zhou, Habermann, Shimada, Golyanik,
  Theobalt, and Xu}{Yi et~al\mbox{.}}{2022}]%
        {PIP}
\bibfield{author}{\bibinfo{person}{Xinyu Yi}, \bibinfo{person}{Yuxiao Zhou},
  \bibinfo{person}{Marc Habermann}, \bibinfo{person}{Soshi Shimada},
  \bibinfo{person}{Vladislav Golyanik}, \bibinfo{person}{Christian Theobalt},
  {and} \bibinfo{person}{Feng Xu}.} \bibinfo{year}{2022}\natexlab{}.
\newblock \showarticletitle{Physical Inertial Poser (PIP): Physics-aware
  Real-time Human Motion Tracking from Sparse Inertial Sensors}. In
  \bibinfo{booktitle}{\emph{IEEE/CVF Conference on Computer Vision and Pattern
  Recognition (CVPR)}}.
\newblock


\bibitem[\protect\citeauthoryear{Yu, Park, and Lee}{Yu et~al\mbox{.}}{2021}]%
        {Yu2021}
\bibfield{author}{\bibinfo{person}{Ri Yu}, \bibinfo{person}{Hwangpil Park},
  {and} \bibinfo{person}{Jehee Lee}.} \bibinfo{year}{2021}\natexlab{}.
\newblock \showarticletitle{Human Dynamics from Monocular Video with Dynamic
  Camera Movements}.
\newblock \bibinfo{journal}{\emph{ACM Trans. Graph.}}  \bibinfo{volume}{40}
  (\bibinfo{year}{2021}).
\newblock


\bibitem[\protect\citeauthoryear{Yuan and Kitani}{Yuan and Kitani}{2019}]%
        {Yuan2019}
\bibfield{author}{\bibinfo{person}{Ye Yuan} {and} \bibinfo{person}{Kris
  Kitani}.} \bibinfo{year}{2019}\natexlab{}.
\newblock \showarticletitle{Ego-Pose Estimation and Forecasting As Real-Time PD
  Control}. In \bibinfo{booktitle}{\emph{2019 IEEE/CVF International Conference
  on Computer Vision (ICCV)}}.
\newblock


\bibitem[\protect\citeauthoryear{Yuan, Wei, Simon, Kitani, and Saragih}{Yuan
  et~al\mbox{.}}{2021}]%
        {SimPoE}
\bibfield{author}{\bibinfo{person}{Ye Yuan}, \bibinfo{person}{Shih-En Wei},
  \bibinfo{person}{Tomas Simon}, \bibinfo{person}{Kris Kitani}, {and}
  \bibinfo{person}{Jason Saragih}.} \bibinfo{year}{2021}\natexlab{}.
\newblock \showarticletitle{SimPoE: Simulated Character Control for 3D Human
  Pose Estimation}. In \bibinfo{booktitle}{\emph{Proceedings of the IEEE/CVF
  Conference on Computer Vision and Pattern Recognition (CVPR)}}.
\newblock


\bibitem[\protect\citeauthoryear{Zell, Rosenhahn, and Wandt}{Zell
  et~al\mbox{.}}{2020}]%
        {Zell2020}
\bibfield{author}{\bibinfo{person}{Petrissa Zell}, \bibinfo{person}{Bodo
  Rosenhahn}, {and} \bibinfo{person}{Bastian Wandt}.}
  \bibinfo{year}{2020}\natexlab{}.
\newblock \showarticletitle{Weakly-supervised Learning of Human Dynamics}. In
  \bibinfo{booktitle}{\emph{ECCV}}.
\newblock


\bibitem[\protect\citeauthoryear{Zhang, Bo, Yong, and Xu}{Zhang
  et~al\mbox{.}}{2019}]%
        {zhang2019interactionfusion}
\bibfield{author}{\bibinfo{person}{Hao Zhang}, \bibinfo{person}{Zi-Hao Bo},
  \bibinfo{person}{Jun-Hai Yong}, {and} \bibinfo{person}{Feng Xu}.}
  \bibinfo{year}{2019}\natexlab{}.
\newblock \showarticletitle{InteractionFusion: real-time reconstruction of hand
  poses and deformable objects in hand-object interactions}.
\newblock \bibinfo{journal}{\emph{ACM Transactions on Graphics (TOG)}}
  \bibinfo{volume}{38}, \bibinfo{number}{4} (\bibinfo{year}{2019}),
  \bibinfo{pages}{1--11}.
\newblock


\bibitem[\protect\citeauthoryear{Zhang, Zhou, Tian, Yong, and Xu}{Zhang
  et~al\mbox{.}}{2021}]%
        {zhang2021single}
\bibfield{author}{\bibinfo{person}{Hao Zhang}, \bibinfo{person}{Yuxiao Zhou},
  \bibinfo{person}{Yifei Tian}, \bibinfo{person}{Jun-Hai Yong}, {and}
  \bibinfo{person}{Feng Xu}.} \bibinfo{year}{2021}\natexlab{}.
\newblock \showarticletitle{Single Depth View Based Real-Time Reconstruction of
  Hand-Object Interactions}.
\newblock \bibinfo{journal}{\emph{ACM Transactions on Graphics (TOG)}}
  \bibinfo{volume}{40}, \bibinfo{number}{3} (\bibinfo{year}{2021}),
  \bibinfo{pages}{1--12}.
\newblock


\bibitem[\protect\citeauthoryear{Zhao, Zhang, Min, and Chai}{Zhao
  et~al\mbox{.}}{2013}]%
        {zhao2013robust}
\bibfield{author}{\bibinfo{person}{Wenping Zhao}, \bibinfo{person}{Jianjie
  Zhang}, \bibinfo{person}{Jianyuan Min}, {and} \bibinfo{person}{Jinxiang
  Chai}.} \bibinfo{year}{2013}\natexlab{}.
\newblock \showarticletitle{Robust realtime physics-based motion control for
  human grasping}.
\newblock \bibinfo{journal}{\emph{ACM Transactions on Graphics (TOG)}}
  \bibinfo{volume}{32}, \bibinfo{number}{6} (\bibinfo{year}{2013}),
  \bibinfo{pages}{1--12}.
\newblock


\end{thebibliography}

\appendix

\end{document}